\def\year{2018}\relax
\documentclass[letterpaper]{article} %DO NOT CHANGE THIS
\usepackage{aaai18}  %Required
\usepackage{times}  %Required
\usepackage{helvet}  %Required
\usepackage{courier}  %Required
\usepackage{url}  %Required
\usepackage{graphicx}  %Required
\begin{document}
% The file aaai.sty is the style file for AAAI Press 
% proceedings, working notes, and technical reports.
%
\title{Learning a Wavelet-like Auto-Encoder to Accelerate Deep Neural Networks}
\author{Tianshui Chen$^1$, Liang Lin$^{1,5}$\thanks{Corresponding author is Liang Lin (Email: linliang@ieee.org). This work was supported by State Key Development Program under Grant 2016YFB1001004, the National Natural
Science Foundation of China under Grant 61622214 and Grant 61320106008, Special Program of the NSFC-Guangdong Joint Fund for Applied Research on Super Computation (the second phase), and Guangdong Natural Science Foundation Project for Research Teams under Grant 2017A030312006.}, Wangmeng Zuo$^2$, Xiaonan Luo$^3$, Lei Zhang$^4$\\
$^1$School of Data and Computer Science, Sun Yat-sen University, Guangzhou, China\\
$^2$School of Computer Science and Technology, Harbin Institute of Technology, Harbin, China\\
$^3$School of Computer Science and Information Security, Guilin University of Electronic Technology, Guilin, China\\
$^4$Department of Computing, The Hong Kong Polytechnic University, HongKong\\
$^5$SenseTime Group Limited \\
}
\maketitle
\begin{abstract}
Accelerating deep neural networks (DNNs) has been attracting increasing attention as it can benefit a wide range of applications, e.g., enabling mobile systems with limited computing resources to own powerful visual recognition ability. A practical strategy to this goal usually relies on a two-stage process: operating on the trained DNNs (e.g., approximating the convolutional filters with tensor decomposition) and fine-tuning the amended network, leading to difficulty in balancing the trade-off between acceleration and maintaining recognition performance. In this work, aiming at a general and comprehensive way for neural network acceleration, we develop a Wavelet-like Auto-Encoder (WAE) that decomposes the original input image into two low-resolution channels (sub-images) and incorporate the WAE into the classification neural networks for joint training. The two decomposed channels, in particular, are encoded to carry the low-frequency information (e.g., image profiles) and high-frequency (e.g., image details or noises), respectively, and enable reconstructing the original input image through the decoding process. Then, we feed the low-frequency channel into a standard classification network such as VGG or ResNet and employ a very lightweight network to fuse with the high-frequency channel to obtain the classification result. Compared to existing DNN acceleration solutions, our framework has the following advantages: i) it is tolerant to any existing convolutional neural networks for classification without amending their structures; ii) the WAE provides an interpretable way to preserve the main components of the input image for classification.
\end{abstract}

\section{Introduction}
Deep convolutional neural networks (CNNs) \cite{lecun1990handwritten,krizhevsky2012imagenet} have been continuously improving the performance of various vision tasks \cite{simonyan2014very,he2016deep,long2015fully,lin2017cross,chen2016disc,chen2016deep}, but at the expense of significantly increased computational complexity. For example, the VGG16-Net \cite{simonyan2014very}, which has demonstrated remarkable performance on a wide range of recognition tasks \cite{long2015fully,ren2015faster,wang2017multi}, requires about 15.36 billion FLOPs\footnote{FLOPs is the number of FLoating-point OPerations} to classify an $224 \times 224$ image. These costs can be prohibitive for the deployment on ordinary personal computers and mobile devices with limited computing resources. Thus, it is highly essential to accelerate the deep CNNs.

\begin{table}[!t]
\centering
\scriptsize
\begin{tabular}{c|c|c|c|c}
\hline
\centering  Methods  & top-5 err.  & CPU (ms) & GPU (ms)  & \#FLOPs\\
\hline
\hline
VGG16 (224)  & 11.65  & 1289.28 & 6.15 & 15.36B \\
VGG16 (112)    &  15.73 & 340.73  (3.78$\times$) &  1.57 (3.92$\times$) & 3.89B\\
Ours   & 11.87  &   411.63 (3.13$\times$) & 2.37 ($2.59\times$) & 4.11B \\
\hline
\end{tabular}
\caption{The top-5 error rate (\%), execution time on CPU and GPU, FLOPs of VGG16-Net with $224 \times 224$ and $112 \times 112$ as input, and our model on the ImageNet dataset. B denotes billion. The error rate is measured on single-view without data augmentation. In the brackets are the acceleration rates compared with the standard VGG16-Net. The execution time is computed with a C++ implementation on Intel i7 CPU (3.50GHz) and Nvidia GeForce GTX TITAN-X GPU.}
\label{table:different_input}
\vspace{-8pt}
\end{table}

There have been a series of efforts dedicated to speed up the deep CNNs, most of which \cite{lebedev2014speeding,tai2015convolutional} employ tensor decomposition to accelerate convolution operations. Typically, these methods conduct two-step optimization separately: approximating the convolution filters of a pre-trained CNN with low-rank decomposition, and then fine-tuning the amended network. This would lead to difficulty in balancing the trade-off between acceleration rate and recognition performance, because two components are not jointly learned to maximize their strengths through cooperation. Another category of algorithms that aim at network acceleration is weight and activation quantization \cite{rastegari2016xnor,cai2017deep,chen2015compressing}, but they usually suffer from an evident drop in performance despite yielding significant speed-up. For example, XNOR-Net \cite{rastegari2016xnor} achieves 58$\times$ speed-up but undergoes 16.0\% top-5 accuracy drop by ResNet-18 on the ImageNet dataset \cite{russakovsky2015imagenet}. Therefore, we present our method according to the following two principles: 1) no explicit network modification such as filter approximation or weight quantitation is needed, which helps to easily generalize to networks with different architectures; 2) the network should enjoy desirable speed-up with tolerable deterioration in performance. 

%As depicted in Table \ref{table:different_input}, down-sampling the input images during both training and testing procedures achieves a significant speed-up but inevitably suffers from a dramatic drop in performance due to the loss of information. 

Since the FLOPs is directly related to the resolution of input images, a seemingly plausible way for acceleration is down-sampling the input images during both training and testing procedures. Although achieving a significant speed-up, it inevitably suffers from a drastic drop in performance due to the loss of information (see Table \ref{table:different_input}). To address this dilemma, we develop a Wavelet-like Auto-Encoder (WAE) that decomposes the original input image into two low-resolution channels and feeds them into the deep CNNs for acceleration. Two decomposed channels are constrained to have following properties: 1) they are encoded to carry low-frequency and high-frequency information, respectively, and are enabled to reconstruct the original image through a decoding process. Thus, most of the content from the original image can be preserved to ensure recognition performance; 2) the high-frequency channel carries minimum information, and thus we can use a lightweight network on it to avoid incurring massive computational burden. In this way, the WAE consists of an encoding layer that decomposes the input image into two channels, and a decoding layer to synthesize the original image based on these two channels. A transform loss, which includes a reconstruction error between the input image and the synthesized image, and an energy minimization loss on the high-frequency channel, is defined to optimize the WAE jointly. Finally, we feed the low-frequency channel to a standard network (e.g., VGG16-Net \cite{simonyan2014very}, ResNet \cite{he2016deep}), and employ a lightweight network to fuse with the high-frequency channel for the classification result.

In the experiments, we first apply our method to the widely used VGG16-Net and conduct extensive evaluations on two large-scale datasets, i.e., the ImageNet dataset for image classification and the CACD dataset for face identification. Our method achieves an acceleration rate of 3.13 with merely 0.22\% top-5 accuracy drop on ImageNet (see Tabel \ref{table:different_input}). On CACD, it even beats the VGG16-Net in performance while achieving the same acceleration rate. Similar experiments with ResNet-50 reveal that even for more compact and deeper network, our method can still achieve 1.88$\times$ speed-up with only 0.8\% top-5 accuracy drop on ImageNet. Note that our method also achieves a better trade-off between accuracy and speed compared with state-of-the-art methods on both VGG16-Net and ResNet. Besides, our method exhibits amazing anti-noise ability compared with the standard network that takes original images as input. Codes are available at \url{https://github.com/tianshuichen/Wavelet-like-Auto-Encoder}.

\section{Related Work}
\noindent\textbf{Tensor decomposition. }Most of the previous works for CNN acceleration focus on approximating the convolution filters by low-rank decomposition \cite{rigamonti2013learning,jaderberg2014speeding,lebedev2014speeding,tai2015convolutional}. As a pioneering work, Rigamonti et al. \cite{rigamonti2013learning} approximate the filters of a pre-trained CNNs with a linear combination of low-rank filters. Jaderberg et al. \cite{jaderberg2014speeding} devise a basis of low-rank filters that are separable in the spatial domain and further develop two different schemes to learn these filters, i.e.,``Filter reconstruction" that minimizes the error of filter weights and ``Data reconstruction" that minimizes the error of the output responses. Lebedev et al. \cite{lebedev2014speeding} adopt a two-stage method that first approximates the convolution kernels using the low-rank CP-decomposition, and then fine-tunes the amended CNN. 
%These methods mainly focused on reducing the inference time. Tai et al. \cite{tai2015convolutional} further proposed to parameterize the convolutional kernels in a way that naturally enforces the low-rank constraint, and it could train the CNN from scratch with low-rank regularization.

\noindent\textbf{Quantization and Pruning. }Weight and activation quantization are widely used for network compression and acceleration. As a representative work, XNOR-Net \cite{rastegari2016xnor} binarizes the input to convolutional layers and filter weights, and approximates convolutions using primarily binary operations, resulting in significant speed-up but an evident drop in performance. Cai et al. \cite{cai2017deep} further introduce Half-Wave Gaussian Quantization to improve the performance of this method. On the other hand, pruning the unimportant connections or filters can also compress and accelerate deep networks. Han et al. \cite{han2015deep} remove the connections with weights below a threshold, reducing the parameters by up to 13$\times$. This method is further combined with weight quantization to achieve an even higher compression rate. Similarly, Li et al. \cite{li2016pruning} measure the importance of a filter by calculating its absolute weight sum and remove the filters with small sum values. Molchanov et al. \cite{molchanov2016pruning} employ the Taylor expansion to approximate the change in the cost function induced by pruning filters. Luo et al. \cite{luo2017thinet} further formulate filter pruning as an optimization problem.

\noindent\textbf{Network structures. }Some works explore more optimal network structures for efficient training and inference.  Lin et al. \cite{lin2013network} develop a low-dimensional embedding method to reduce the number and size of the filters. Simonyan et al. \cite{simonyan2014very} show that stacked filters with small spatial dimensions (e.g., $3 \times 3$) could operate in the same receptive field of larger filters (e.g., $5 \times 5$) with less computational complexity. Iandola et al. \cite{iandola2016squeezenet} further replace some $3 \times 3$ filters with $1 \times 1$ filters, and decrease the number of input channels to $3 \times 3$ filters to simultaneously speed up and compress the deep networks. 

Different from the aforementioned methods, we learn a WAE that decomposes the input image into two low-resolution channels and utilizes these decomposed channels as inputs to the CNN to reduce the computational complexity without compromising accuracy. Compared with existing methods, our method does not amend the network structures, and thus it can easily generalize to any existing convolutional neural networks.

%It can be regarded as a complementary method of the existing methods, and has notable potential to work cooperatively with them to further improve the CNN efficiency.

\begin{figure*}[htbp]
   \centering
   \includegraphics[width=0.75\linewidth]{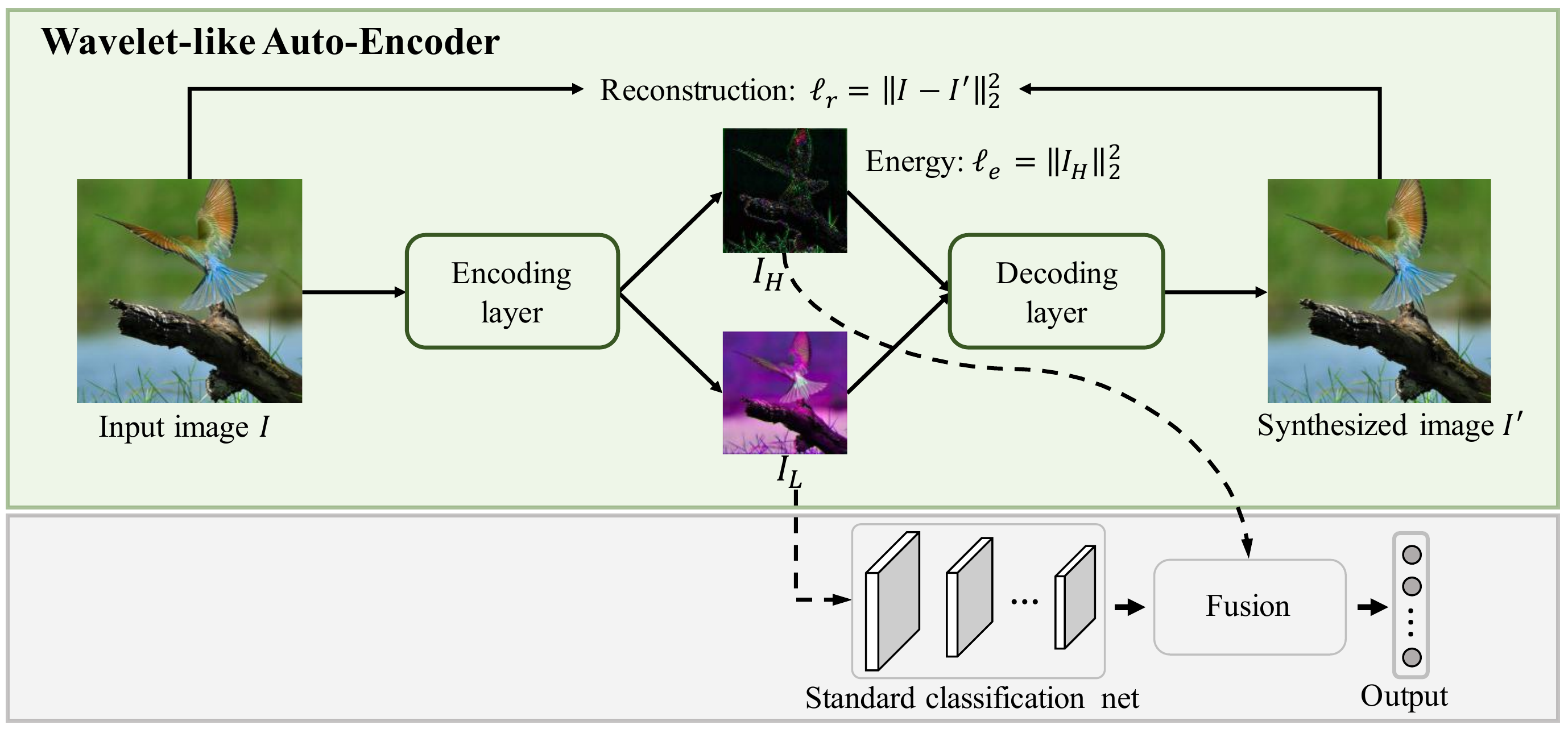}
   \vspace{-6pt}
   \caption{The overall framework of our proposed method. The key component of the framework is the WAE that decomposes an input image into two low-resolution channels, i.e., $I_L$ and $I_H$. These two channels encode the high- and low-frequency information respectively and are enabled to construct the original image via a decoding process. The low-frequency channel is then fed into the a standard network (e.g., VGG16-Net or ResNet) to extract its features. Then a lightweight network fuses these features and the high-frequency channel to predict the label scores. Note that the input to the classification network is low-resolution; thus it enjoys higher efficiency.}
   \label{fig:cls_net}
\end{figure*}

\section{Proposed Method}
The overall framework of our proposed method is illustrated in Figure \ref{fig:cls_net}. The WAE decomposes the input image into two low-resolution channels that carry low-frequency information (e.g., basic profile) and high-frequency (e.g., auxiliary details), i.e., $I_L$ and $I_H$, and these two channels are enabled to construct the original image through the decoding process. Finally, the low-frequency channel is fed into a standard network (e.g., VGG16-Net or ResNet) to produce its features, and a network is further employed to fuse these features with the high-frequency channel to predict the classification result.

\subsection{Wavelet-like Auto-Encoder}
The WAE consists of an encoding layer that decomposes the input image into two low-resolution channels and a decoding layer that synthesizes the original input image based on these two decomposed channels. In the following context, we introduce the image decomposition and synthesis processes in detail.

\begin{figure}[!t]
   \centering
   \includegraphics[width=0.7\linewidth]{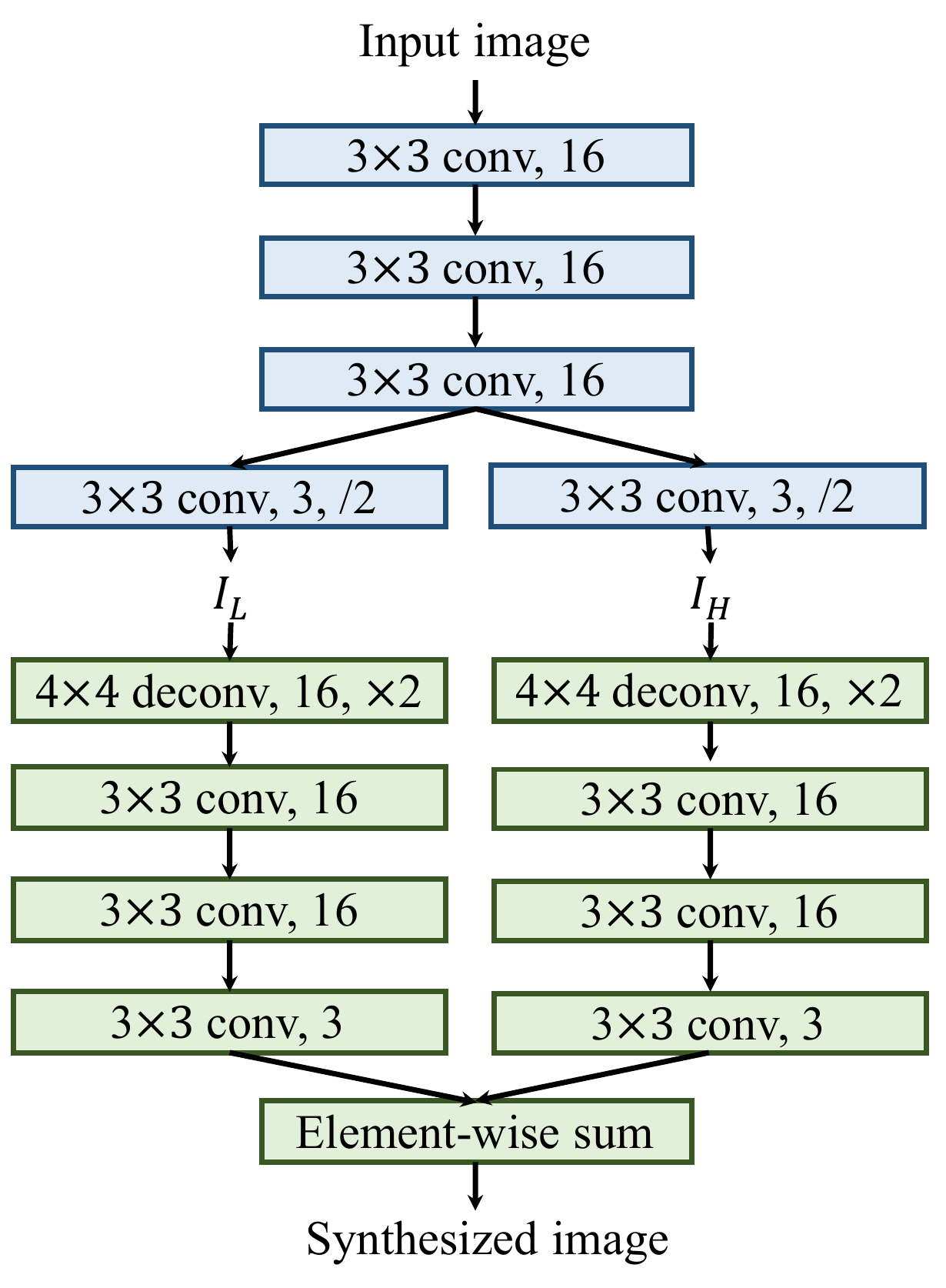}
   \vspace{-8pt}
   \caption{Detailed architecture of the wavelet-like auto-encoder. It consists of an encoding (the blue part) and a decoding (the green part) layers. ``/2" denotes a conv layer with a stride of 2 to downsample the feature maps, and conversely ``$\times 2$" denotes a deconv layer with a stride of 2 to upsample the feature maps.}
   \label{fig:dr_net}
      \vspace{-8pt}
\end{figure}

\noindent\textbf{Image decomposition}. Given an input image $I$ of size $W\times H$, it is first decomposed into two half-resolution channels, i.e., $I_L$ and $I_H$, which is formulated as:
\begin{equation} 
   [I_L, I_H] = \mathcal{F}_E(I, \mathbf{W}_E),
\end{equation}
where $\mathcal{F}_E$ denotes the encoding process, and $\mathbf{W}_E$ are its parameters. In this paper, the encoding layer contains three stacked convolutional (conv) layers with strides of 1, followed by two branched conv layers with strides of 2 to produce $I_L$ and $I_H$, respectively. It is intolerable if this process incurs massive computational overhead, as we focus on acceleration. To ensure efficient computing, we utilize the small kernels with sizes of $3 \times 3$ and set the channel numbers of all intermediate layers as 16. The detailed architecture of the encoding layer are illustrated in Figure \ref{fig:dr_net} (the blue part).

\noindent\textbf{Image synthesis}. 
The decoding layer is employed to synthesize the input image based on $I_L$ and $I_H$. It processes $I_L$ and $I_H$ to get the up-sampled images $I{'}_L$ and $I{'}_H$, separately, and then simply adds them to obtain the synthesized image $I{'}$. The process is formulated as:
\begin{equation} 
   I' = \mathcal{F}_{D_L}(I_L, \mathbf{W}_{D_L}) + \mathcal{F}_{D_H}(I_H, \mathbf{W}_{D_H}),
\end{equation}
where $\mathcal{F}_{D_L}$ and $\mathcal{F}_{D_H}$ are the transforms on $I_L$ and $I_H$, and  $\mathbf{W}_{D_L}$ and $\mathbf{W}_{D_H}$ are their parameters. The decoding layer has two branches that share the same architecture, with each branch implementing one transform. Each branch contains a deconvolutional (deconv) layer with a stride of 2 and three stacked conv layers with strides of 1, in which the deconv layer first up-samples the input images (i.e., $I_L$ and $I_H$) by two times, and the conv layers further refine the up-sampled feature maps to generate the outputs (i.e., $I{'}_L$ and $I{'}_H$). We set the kernel size of the deconv layer as $4 \times 4$, and those of the conv layers as $3 \times 3$. The channel numbers of all the intermediate layers are also set as 16. We present the detailed architecture of the decoding layer in Figure \ref{fig:dr_net} (the green part).

\subsection{Encoding and decoding constraints}
We adopt the two decomposed channels $I_L$ and $I_H$ to replace the original image as input to the classification network.  To ensure the classification performance and simultaneously consider the computational overhead, $I_L$ and $I_H$ are expected to possess two properties as follows:
\begin{itemize}
  \item \textbf{Minimum information loss. } $I_L$ and $I_H$ should retain all the content of the original image as the classification network does not see the original image directly. If some discriminative content is lost unexpectedly, it may lead to classification error.
  \item \textbf{Minimum $I_H$ energy. } $I_H$ should contain minimum information, so we can apply a lightweight network to it to avoid incurring heavy computational overhead.
\end{itemize}
To comprehensively consider these two properties, we define a transform loss that consists of two simple yet effective constraints on the decomposed and reconstructed results. 

\noindent\textbf{Reconstruction constraint}. 
An intuitive assumption is that if $I_L$ and $I_H$ preserve all the content of the input image, they are enabled to construct the input image. In this paper, we reconstruct image $I'$ from $I_L$ and $I_H$ through the decoding process, and minimize the reconstruction error between input image $I$ and reconstructed image $I'$. So the reconstruction constraint can be formulated as:

\begin{equation} 
   \ell_r=\vert\vert I-I' \vert\vert_2^2.
\end{equation}

\noindent\textbf{Energy constraint}. 
The second constraint minimizes the energy of $I_H$ and pushes most information to $I_L$ thus that $I_H$ preserves minimum content. It can be formulated as: 

\begin{equation} 
   \ell_e=\vert\vert I_H \vert\vert_2^2.
\end{equation}
We combine the two constraints on the decomposed and reconstructed results to define a transform loss. It is formulated as the weighted sum of these two constraints:

\begin{equation} 
   \mathcal{L}_t=\ell_r+\lambda\ell_e,
\end{equation}
where $\lambda$ is a weighted parameter, and it is set as 1 in our experiments.

\subsection{Image classification}
The image classification network consists of a standard network to extract the features $f_L$ for $I_L$, and a fusion network to fuse with $I_H$ to predict the final label scores. Here, the standard network refers to the VGG16-Net or the ResNet, and we use the features maps from the last conv layer. The fusion network contains a sub-module to extract features $f_H$ for $I_H$. The sub-module shares the same architecture with the standard network except that the channel numbers of all the conv layers are divided by 4. $f_L$ is fed into a simple classifier to predict a score vector $\mathbf{s}_L$, and it is further concatenated with $f_H$ to compute a score vector $\mathbf{s}_c$ by a similar classifier. The two vectors are then averaged to obtain the final score vector $\mathbf{s}$.

We employ the cross entropy loss as our objective function to train the classification network. Suppose there are $N$ training samples, and each sample $I_i$ is annotated with a label $y_i$. 
Given the predicted probability vector $\mathbf{p}_i$
 \begin{equation}
      p_{i}^c= \frac{\exp(s_{i}^c)}{\sum_{c'=0}^{C-1}\exp(s_{i}^{c'})} \ c=0,1,\dots,C-1,
\end{equation}
where $C$ is the number of class labels. The classification loss function is expressed as:
 \begin{equation}
      \mathcal{L}_{{1}}=-\frac{1}{N}\sum_{i=1}^N\sum_{c=0}^{C-1}\mathbf{1}(y_i=c)\log{p_c},
\end{equation}
where $\mathbf{1}(\cdot)$ is the indicator function whose value is 1 when the expression is true, and 0 otherwise. We use the same loss function for $\mathbf{s}_c$ and $\mathbf{s}_L$ and simply sum up them to get the final classification loss $\mathcal{L}_c$. 

\noindent\textbf{Discussion on computational complexity. } As suggested in \cite{he2015convolutional}, the convolutional layers often take 90-95\% computational cost. Here, we analyze the computational complexity of the convolutional layers and present an up bound of the acceleration rate of our method compared with the standard network. For a given CNN, the total complexity of all the convolutional layers can be expressed as:  
\begin{equation} 
   \mathcal{O}(\sum_{l=1}^{d}n_{l-1} \cdot s_l^2 \cdot n_l \cdot m_l^2),
\end{equation}
where $d$ is the number of the conv layers, and $l$ is the index of the conv layer; $n_l$ is the channel number of the $l$-th layer; $s_l$ is the spatial size of the kernels and $m_l$ is the spatial size of the output feature maps. For the standard network, as it takes a half-resolution image as input, $m_l$ of each layer is also halved. So the computational complexity is about $\frac{1}{4}$ of the standard network that takes original images as input. For the sub-module in fusion network that processes $I_H$, the channel number of all the corresponding layers are further quartered, so the computational complexity is merely about $\frac{1}{64}$ of the original standard network. So the up bound of the acceleration rate of our classification network compared with the standard network can be estimated by $\frac{1} {1/4 + 1/64}=3.76$. However, as the decomposition process and additional fully-connected layers could incur additional overhead, the actual acceleration rate may be lower than 3.76.

\subsection{Learning}
Our model is comprised of the WAE and the classification network, and it is indeed possible to jointly train them using a combination of the transform and classification losses in an end-to-end manner. However, it is difficult to balance the two loss terms if directly training from scratch, inevitably leading to inferior performance. To address this issue, the training process is empirically divided into three stages:

\noindent\textbf{Stage 1: WAE training}. We first remove the classification network and train the WAE using the transform loss $\mathcal{L}_t$. Given an image, we first resize it to $256 \times 256$ and randomly extract patches (and their horizontal reflections) with a size of $224 \times 224$, and train the network based on these extracted patches. The parameters are  initialized with the Xavier algorithm \cite{glorot2010understanding} and the WAE is trained using SGD algorithm with a mini-batch of 4, momentum of 0.9 and weight decay of 0.0005. We set the initial learning rate as 0.000001, and divide it by 10 after 10 epochs.

\noindent\textbf{Stage 2: Classification network training}. We combine the classification network with the WAE, and train the classification network with the classification loss $\mathcal{L}_{{c}}$. The parameters of the WAE are initialized with the parameters learned in Stage 1 and are kept fixed, and the parameters of classification network are also initialized with the Xavier algorithm. The training images are resized to $256 \times 256$, and the same strategies (i.e., random cropping and horizontal flipping) are adopted for data augmentation. The network is also trained using SGD algorithm with the same momentum and weight decay as Stage 1. The mini-batch is set as 256, and the learning rate is initialized as 0.01 (0.1 if using ResNet-50 as the baseline), which is divided by 10 when the error plateaus.

\noindent\textbf{Stage 3: Joint fine tuning.} To better adapt the decomposed channels for classification, we also fine tune WAE and classification network jointly by combining the transform and classification losses, formulated as:
 \begin{equation}
      \mathcal{L}=\mathcal{L}_c+\gamma\mathcal{L}_t,
\end{equation}
where $\gamma$ is set to be 0.001 to balance the two losses. The network is fine tuned using SGD with the mini-batch size, momentum, and weight decay the same as Stage 2. We utilize a small learning rate of 0.0001 and train the network until the error plateaus.

\section{Experiments}
\subsection{Baseline methods}
In the experiments, we utilize two popular standard networks, i.e., VGG16-Net and ResNet-50, as the baseline networks, and mainly compare with these baselines on image recognition performance and execution efficiency. To further validate the effectiveness of the proposed WAE, we implement two baseline methods that also utilize the decomposed channels as input to the deep CNNs for classification. 

\noindent\textbf{Wavelet+CNN. }Discrete wavelet transforms (DWTs) decompose an input image to four half-resolution channels, i.e., cA, cH, cV, cD, where cA is an approximation to the input image (similar to $I_L$), while cH, cV, cD preserve image details (similar to $I_H$). Also, the original image can be reconstructed based on cA, cH, cV, cD using an inverse transform. Then, cA is fed into the standard network, and cH, cV, cD is concatenated and fed into the final for classification. Here, we use the widely used 9/7 implementation \cite{zhang2011fsim} for the DWT.

\noindent\textbf{Decomposition+CNN. }This method also uses an encoding layer the same to that in our proposed WAE to decompose the input image into two half-resolution ones, followed by the classification network for predicting the class labels. But it has no constraints on the decomposed channels, and it is trained merely with the classification loss.

\begin{table*}[!t]
\centering
\begin{tabular}{c|c|c|c|c|c}
\hline
\centering  Methods  & top-5 err. (\%)& CPU (ms)  & CPU speed-up rate  & GPU (ms) & GPU speed-up rate\\
\hline
\hline
VGG16-Net   & 11.65  & 1289.28& 1$\times$ & 6.15&1$\times$  \\
\hline
Wavelet+CNN      &  14.42 & 392.24  & $3.29\times$  & 2.30  & 2.67$\times$  \\
Decomposition+CNN      &  12.98 & 411.63 & 3.13$\times$ & 2.37  & $2.59\times$\\
\hline
Taylor-1 & 13.00  & - & 1.70$\times$ & - & 2.20$\times$ \\
Taylor-2 & 15.50  & - & 2.10$\times$ & - & 3.40$\times$ \\
ThiNet-Tiny & 18.03  & 116.25  & 11.25$\times$ & 1.32 & 4.66$\times$ \\
ThiNet-GAP & 12.08  &442.807  & 2.91$\times$ &2.52 & 2.44$\times$\\
\hline
Ours   & 11.87  & 411.63 & 3.13$\times$ & 2.37  & 2.59$\times$ \\
\hline
\end{tabular}
\vspace{-8pt}
\caption{Comparison of the top-5 error rate, execution time and speed-up rate on CPU and GPU of VGG16-Net, the two baseline methods and the previous state of the art methods on the ImageNet dataset. The error rate is measured on single-view without data augmentation.}
\label{table:imagenet_comparison}
\end{table*}

The classification networks in the two baseline methods share the same structure with that of ours for fair comparisons. Both two methods are also trained with SGD with an initial learning rate of 0.01, mini-batch of 256, momentum of 0.9 and weight decay of 0.0005. We select the models with lowest validation errors for comparison.

\subsection{ImageNet classification with VGG16}
We first evaluate our proposed method on VGG16-Net on the ImageNet dataset \cite{russakovsky2015imagenet}. The dataset covers 1,000 classes and comprises a training set of about 1.28 million images and a validation set of 50,000 images. All the methods are trained on the training set, and evaluated on the validation set as the ground truth of the test set are not available. Following \cite{zhang2016accelerating}, we report the top-5 error rate for performance comparison.

\subsubsection{Comparison with the original VGG16-Net}
We first compare the classification performance and execution time on CPU and GPU of our model and the original VGG16-Net\footnote{For a fair comparison, the top-5 error rate of the original VGG16-Net is evaluated with center-cropped patches on resized images. The same strategy is also used in ResNet-50.} in Table \ref{table:imagenet_comparison}. The execution time is evaluated with a C++ implementation on Intel i7 CPU (3.50GHz) and Nvidia GeForce GTX TITAN-X GPU. We can see our model achieves a speed-up rate of up to $3.13\times$ with merely 0.22\% increase in the top-5 error rate. For the CPU version, our model obtains an actual acceleration rate of $3.13\times$, close to the up bound of the acceleration rate ($3.76\times$). The overhead may come from the computational cost of the encoding layer and additional fully-connected layers. For the GPU version, the actual acceleration rate is $2.59\times$. It is smaller since an accelerated model is harder for parallel computing.

\subsubsection{Comparison with state-of-the-art methods}
ThiNet \cite{luo2017thinet} and Taylor\footnote{The execution time reported in Taylor paper are conducted on a hardware and software platform that is different from ours. Thus we merely present the relative speed-up rates for fair comparison.} \cite{molchanov2016pruning} are two newest methods that also focus on accelerating deep CNNs, and they also conduct experiments on VGG16-Net. In this part, we compare our model with these methods and report the results in Table \ref{table:imagenet_comparison}. Taylor presents two models, namely Taylor-1 and Taylor-2. Our model achieves better accuracy and speed-up rate than Taylor-1. The speed-up rate of Taylor-2 is a bit higher than ours, but it suffers an evident performance drop (3.63\% increase in top-5 error rate). ThiNet also presents two models, i.e., ThiNet-GAP and ThiNet-Tiny. ThiNet-Tiny enjoys a significant speed-up at the cost of a drop in accuracy (6.47\% increase in top-5 error rate), which is intolerant for real-world systems. ThiNet-GAP can achieve a better trade-off between speed and accuracy, but our model still surpasses it in both speed and accuracy. 

\begin{figure}[htbp]
   \centering
   \includegraphics[width=0.90\linewidth]{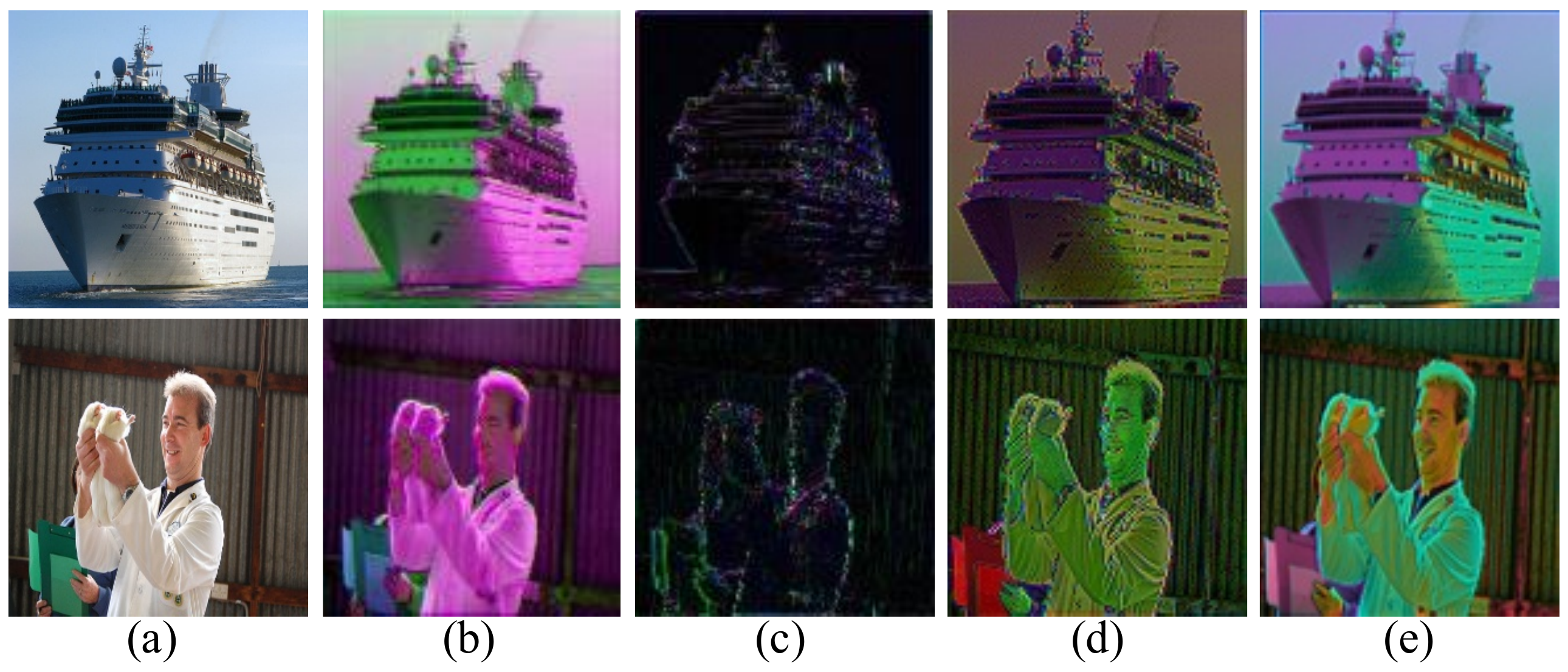}
   \caption{Visualization results of the input image (a), the sub-images (b) and (c) produced by our method and the sub-images (d) and (e) produced by the ``Decomposition+CNN''.}
   \label{fig:visualization_comparison}
\end{figure}

\subsubsection{Comparison with the baseline methods}
To validate the effectiveness of the proposed WAE, we also compare our model with two baseline methods that use different decomposition strategies in Table \ref{table:imagenet_comparison}. It shows our model outperforms the two baseline methods by a sizable margin on the classification performance while sharing comparable efficiency. Note that ``Wavelet+CNN" runs a bit faster than our method, as it uses the more efficient DWT for image decomposition. However, it results in inferior performance, and one possible reason is that directly separating the low- and high-frequency information of an image may hamper the classification result. Our model also decomposes the input image into two channels, but it pushes most information to the $I_L$ via minimizing the energy of $I_H$ and are jointly trained to better adapt for classification. We will conduct experiments to analyze the classification performance merely using $I_L$ and cA to give a deeper comparison later. To compare the difference between our model and the ``Decomposition+CNN", we visualize the decomposed channels generated by this method and ours in Figure \ref{fig:visualization_comparison}. Without the constraints, the two decomposed channels share identical appearance, and fusing the classification results of them can be regarded as the model ensemble. Conversely, the channels generated by our model are somehow complementary, as $I_L$ retains the main content, while $I_H$ preserves the subtle details. These comparisons well prove the proposed WAE can achieve a better balance between speed and accuracy.

\begin{figure}[!t]
   \centering
   \includegraphics[width=0.90\linewidth]{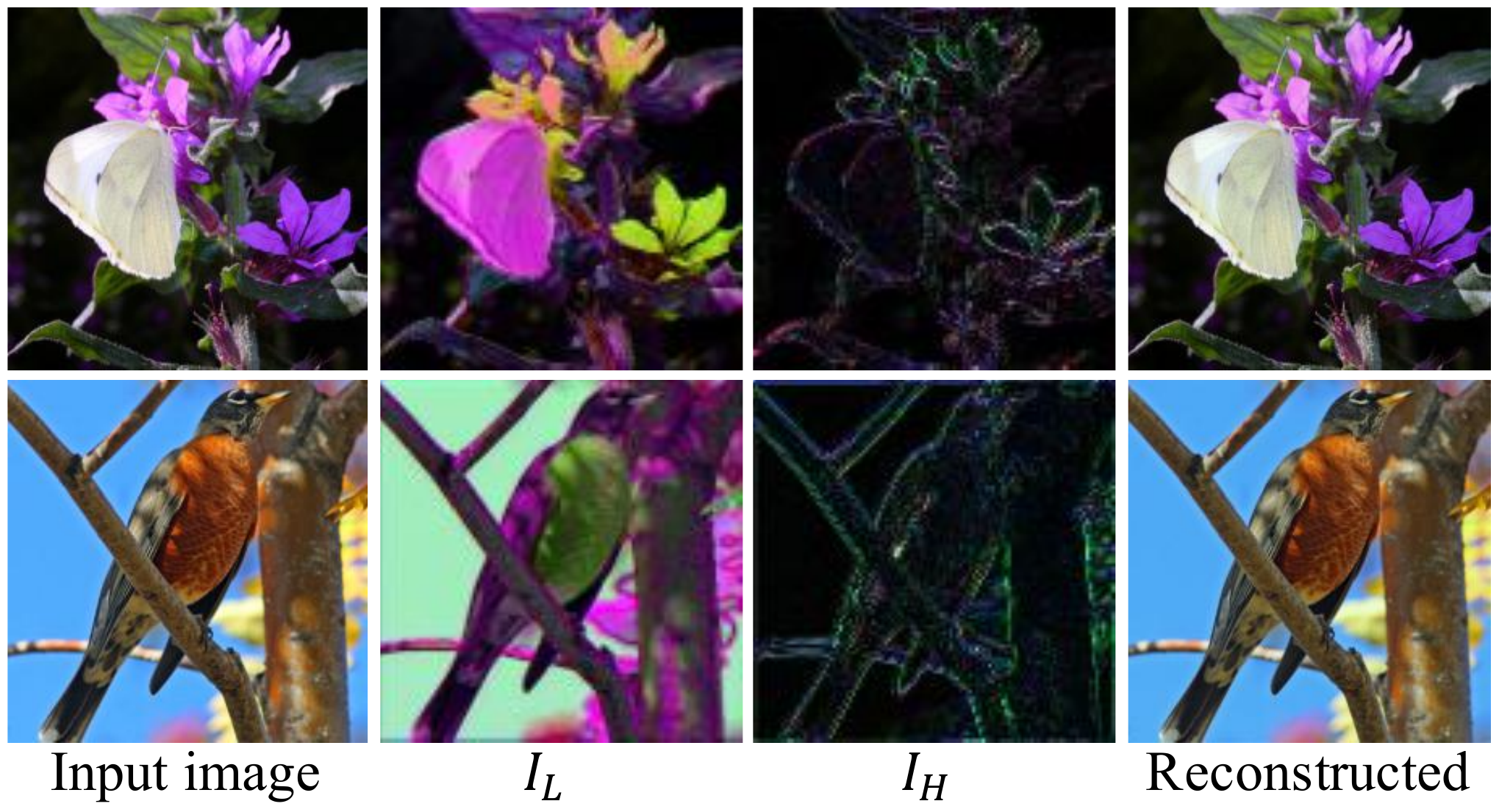}
   \caption{Visualization results of the $I_L$, $I_H$ and the reconstructed image.}
   \label{fig:decomposition_visualization}
\end{figure}

\subsubsection{Analysis on the decomposed channels}
Some examples of the decomposed channels and the reconstructed images are visualized in Figure \ref{fig:decomposition_visualization}. We can observe that $I_L$ indeed contains the main content of the input image, while $I_H$ preserves the details, e.g., edges and contours. It also shows excellent reconstructed results. These visualization results finely accord with the assumption of our method. 

\begin{table}[htbp]
\centering
\begin{tabular}{c|c|c|c|c}
\hline
\centering  Input  & cA & $I_R$ & $I_L$    & $I_L$+$I_H$ \\
\hline
\hline
\centering top-5 err.  & 15.92 & 15.73  & 14.20 & 11.87 \\
\hline
\end{tabular}
\caption{Comparison of the top-5 error rate using $I_L$+$I_H$, $I_L$, $I_R$ and cA for classification on the ImageNet dataset.}
\label{table:I_L}
\end{table}

To provide deeper analysis of the decomposed channels, we present the performance using the $I_L$ for classification. We first exclude the fusing with $I_H$ and re-train the classification network with the parameters of WAE fixed. The top-5 error rate is depicted in Table \ref{table:I_L}. It is not surprising that the performance drops, as $I_H$ preserves the image details and can provide auxiliary information for classification. We also conduct experiments that use cA generated by DWT, and $I_R$ generated by directly resizing the image to a given size, for classification. Specifically, we first resize cA and $I_R$ to $128 \times 128$, and randomly crop the patches of size $112 \times 112$ (and their horizontal reflections) for training. During testing, we crop the center patch with a size of $112 \times 112$ for fair comparisons. Although $I_L$, $I_R$ and cA are all assumed to possess the main content of the original image, the classification result using $I_L$ is obviously superior to those using $I_R$ and cA. One possible reason is that the constraint on minimizing the energy of $I_H$ explicitly pushes most content to $I_L$ so that $I_L$ contains much more discriminative information than $I_R$ and cA. These comparisons can also give a possible explanation that our approach outperforms the ``Wavelet+CNN". 

%\begin{table}[htbp]
%\centering
%\begin{tabular}{c|c}
%\hline
%\centering  Methods  & top-5 err. (\%)\\
%\hline
%\hline
%cA    & 15.92   \\
%$I_R$    & 15.73   \\
%$I_L$    & 14.20   \\
%$I_L$+$I_H$ & 11.87 \\
%\hline
%\end{tabular}
%\vspace{4pt}
%\caption{Comparison of the top-5 error rate using $I_L$+$I_H$, $I_L$, $I_R$ and cA for classification on the ImageNet dataset.}
%\label{table:I_L}
%\end{table}

\subsubsection{Contribution of joint fine tuning step}
We evaluated the contribution of joint fine tuning by comparing the performance with and without it, as reported in Table \ref{table:ft}. The top-5 error rates with fine tuning decreases by 0.34\%. This suggests fine tuning the network jointly can adapt the decomposed image for better classification.

\begin{table}[htbp]
\centering
\begin{tabular}{c|c|c}
\hline
\centering  Methods  & w/o FT & w/ FT\\
\hline
\hline
top-5 err. (\%) & 12.21 & 11.87 \\
\hline
\end{tabular}
\caption{Comparison of the top-5 error rate with and without joint fine tuning (FT) on the ImageNet dataset.}
\label{table:ft}
\end{table}

\subsection{ImageNet classification with ResNet-50}
In this part, we further evaluate the performance of our proposed method on ResNet. Without loss of generalization, we select ResNet-50 from the ResNet family and simply use it to replace the VGG-Net as the baseline network. Then it is trained from scratch using a similar process as described in the Sec. of Learning. Because ResNet is a recently proposed network architecture, few works are proposed to accelerate this network. Thus, we simply compared with the standard ResNet-50, ThiNet in Table \ref{result:reset}. ResNet is a more compact model, and accelerating this network is even more difficult. However, our method can still achieve 1.88$\times$ speed-up with merely 0.8\% increase in top-5 error rate, surpassing ThiNet on both accuracy and efficiency.

\begin{table}[htbp]
\centering
\begin{tabular}{c|c|c|c}
\hline
\centering  Methods  & top-5 err. (\%) & GPU SR & CPU SR\\
\hline
\hline
ResNet-50   & 8.86 & 1$\times$  & 1$\times$ \\
ThiNet-30    & 11.70 & 1.30$\times$  & -    \\
Ours & 9.66 & 1.73$\times$ & 1.88$\times$ \\
\hline
\end{tabular}
\caption{Comparison of the top-5 error rate and speed-up rate (SR) of our model and ThiNet on ResNet-50 on the ImageNet dataset.}
\label{result:reset}
\end{table}

\subsection{CACD face identification}
CACD is a large-scale and challenging dataset for face identification. It contains 163,446 images of 2,000 identities collected from the Internet that vary in age, pose and illumination. A subset of 56,138 images that cover 500 identities are manually annotated \cite{lin2017active}. We randomly select 44,798 images as the training set and the rest as the test set. All the models are trained on the training set and evaluated on the test set. Table \ref{table:cacd_comparison} presents the comparison results. Note that the execution times are the same as Table \ref{table:imagenet_comparison}. In this dataset, our model outperforms the VGG16-Net (0.22\% increase in accuracy) and meanwhile achieves a speed-up rate of $3.13\times$. Besides, our method also beats the baseline methods. These comparisons again demonstrate the superiority of our proposed WAE. Remarkably, the images on CACD are far different from those on ImageNet, and our method still achieves superior performance on both accuracy and efficiency. It suggests our model can generalize to diverse datasets for accelerating the deep CNNs.

\begin{table}[htbp]
\centering
\begin{tabular}{c|c}
\hline
\centering  Methods  & acc. (\%) \\
\hline
\hline
VGG16-Net    & 95.91   \\
Wavelet+CNN      &   94.99  \\
Decomposition+CNN      &   95.20 \\
Ours   & 96.13   \\
\hline
\end{tabular}
\caption{Comparison of the accuracy of our model, VGG16-Net and the baseline methods on the CACD dataset.}
\label{table:cacd_comparison}
\end{table}

\subsection{Noisy image classification}
Generally, the high-frequency part of an image contains more noise. Our model may implicitly remove some high-frequency part by minimize the energy of $I_H$, so it may be inherently more robust to the noise. To validate this assumption, we add Gaussian noise of mean zero and different variances V to the test images, and present the accuracy of our method and the original VGG16-Net on these noisy images in Table \ref{table:cacd_noise}. Note that both our model and the VGG16-Net is trained with the clean images. Our model performs consistently better than VGG16-Net over different noise levels. Remarkably, the superiority of our model is more evident when adding larger noise. For example, when adding noise with a variance of 0.05, our model outperforms the VGG16-Net by 10.81\% in accuracy. These comparisons suggest our method is more robust to noise compared to VGG16-Net.

\begin{table}[htbp]
\centering
\begin{tabular}{c|c|c}
\hline
Methods & VGG16-Net & Ours \\
\hline
V=0   & 95.91 & 96.13    \\
V=0.01   & 90.22  &  91.16     \\
V=0.02   & 80.00  &  83.85     \\
V=0.05   & 45.10  &  55.91     \\
V=0.1   & 14.31  &   23.88    \\
\hline
\end{tabular}
\caption{Comparison of accuracy (in \%) on the image of our model and VGG16-Net with gaussian noise of zero mean and different variances on the CACD dataset.}
\label{table:cacd_noise}
\end{table}

\section{Conclusion}
In this paper, we learn a Wavelet-like Auto-Encoder, which decomposes an input image into two low-resolution channels and utilizes the decomposed channels as inputs to the CNN to reduce the computational complexity without compromising the accuracy. Specifically, the WAE consists of an encoding layer to decompose the input image into two half-resolution channels and a decoding layer to synthesize the original image from the two decomposed channels. A transform loss, which combines a reconstruction error that constrains the two low-resolution channels to preserve all the information of the input image, and an energy minimization loss that constrain one channel contains minimum energy, are further proposed to optimize the network. 
In future work, we will conduct experiments to decompose the image into sub-images of lower resolution to explore a better trade-off between accuracy and speed.

\bibliographystyle{aaai}
\bibliography{reference}

\begin{thebibliography}{}

\bibitem[\protect\citeauthoryear{Cai \bgroup et al\mbox.\egroup
  }{2017}]{cai2017deep}
Cai, Z.; He, X.; Sun, J.; and Vasconcelos, N.
\newblock 2017.
\newblock Deep learning with low precision by half-wave gaussian quantization.
\newblock {\em arXiv preprint arXiv:1702.00953}.

\bibitem[\protect\citeauthoryear{Chen \bgroup et al\mbox.\egroup
  }{2015}]{chen2015compressing}
Chen, W.; Wilson, J.; Tyree, S.; Weinberger, K.; and Chen, Y.
\newblock 2015.
\newblock Compressing neural networks with the hashing trick.
\newblock In {\em International Conference on Machine Learning},  2285--2294.

\bibitem[\protect\citeauthoryear{Chen \bgroup et al\mbox.\egroup
  }{2016}]{chen2016disc}
Chen, T.; Lin, L.; Liu, L.; Luo, X.; and Li, X.
\newblock 2016.
\newblock Disc: Deep image saliency computing via progressive representation
  learning.
\newblock {\em IEEE transactions on neural networks and learning systems}
  27(6):1135--1149.

\bibitem[\protect\citeauthoryear{Chen, Guo, and Lai}{2016}]{chen2016deep}
Chen, S.-Z.; Guo, C.-C.; and Lai, J.-H.
\newblock 2016.
\newblock Deep ranking for person re-identification via joint representation
  learning.
\newblock {\em IEEE Transactions on Image Processing} 25(5):2353--2367.

\bibitem[\protect\citeauthoryear{Glorot and
  Bengio}{2010}]{glorot2010understanding}
Glorot, X., and Bengio, Y.
\newblock 2010.
\newblock Understanding the difficulty of training deep feedforward neural
  networks.
\newblock In {\em Aistats}, volume~9,  249--256.

\bibitem[\protect\citeauthoryear{Han, Mao, and Dally}{2015}]{han2015deep}
Han, S.; Mao, H.; and Dally, W.~J.
\newblock 2015.
\newblock Deep compression: Compressing deep neural networks with pruning,
  trained quantization and huffman coding.
\newblock {\em arXiv preprint arXiv:1510.00149}.

\bibitem[\protect\citeauthoryear{He and Sun}{2015}]{he2015convolutional}
He, K., and Sun, J.
\newblock 2015.
\newblock Convolutional neural networks at constrained time cost.
\newblock In {\em Proceedings of the IEEE Conference on Computer Vision and
  Pattern Recognition},  5353--5360.

\bibitem[\protect\citeauthoryear{He \bgroup et al\mbox.\egroup
  }{2016}]{he2016deep}
He, K.; Zhang, X.; Ren, S.; and Sun, J.
\newblock 2016.
\newblock Deep residual learning for image recognition.
\newblock In {\em Proceedings of the IEEE Conference on Computer Vision and
  Pattern Recognition},  770--778.

\bibitem[\protect\citeauthoryear{Iandola \bgroup et al\mbox.\egroup
  }{2016}]{iandola2016squeezenet}
Iandola, F.~N.; Han, S.; Moskewicz, M.~W.; Ashraf, K.; Dally, W.~J.; and
  Keutzer, K.
\newblock 2016.
\newblock Squeezenet: Alexnet-level accuracy with 50x fewer parameters and< 0.5
  mb model size.
\newblock {\em arXiv preprint arXiv:1602.07360}.

\bibitem[\protect\citeauthoryear{Jaderberg, Vedaldi, and
  Zisserman}{2014}]{jaderberg2014speeding}
Jaderberg, M.; Vedaldi, A.; and Zisserman, A.
\newblock 2014.
\newblock Speeding up convolutional neural networks with low rank expansions.
\newblock {\em arXiv preprint arXiv:1405.3866}.

\bibitem[\protect\citeauthoryear{Krizhevsky, Sutskever, and
  Hinton}{2012}]{krizhevsky2012imagenet}
Krizhevsky, A.; Sutskever, I.; and Hinton, G.~E.
\newblock 2012.
\newblock Imagenet classification with deep convolutional neural networks.
\newblock In {\em Advances in neural information processing systems},
  1097--1105.

\bibitem[\protect\citeauthoryear{Lebedev \bgroup et al\mbox.\egroup
  }{2014}]{lebedev2014speeding}
Lebedev, V.; Ganin, Y.; Rakhuba, M.; Oseledets, I.; and Lempitsky, V.
\newblock 2014.
\newblock Speeding-up convolutional neural networks using fine-tuned
  cp-decomposition.
\newblock {\em arXiv preprint arXiv:1412.6553}.

\bibitem[\protect\citeauthoryear{LeCun \bgroup et al\mbox.\egroup
  }{1990}]{lecun1990handwritten}
LeCun, Y.; Boser, B.~E.; Denker, J.~S.; Henderson, D.; Howard, R.~E.; Hubbard,
  W.~E.; and Jackel, L.~D.
\newblock 1990.
\newblock Handwritten digit recognition with a back-propagation network.
\newblock In {\em Advances in neural information processing systems},
  396--404.

\bibitem[\protect\citeauthoryear{Li \bgroup et al\mbox.\egroup
  }{2016}]{li2016pruning}
Li, H.; Kadav, A.; Durdanovic, I.; Samet, H.; and Graf, H.~P.
\newblock 2016.
\newblock Pruning filters for efficient convnets.
\newblock {\em arXiv preprint arXiv:1608.08710}.

\bibitem[\protect\citeauthoryear{Lin \bgroup et al\mbox.\egroup
  }{2017a}]{lin2017cross}
Lin, L.; Wang, G.; Zuo, W.; Feng, X.; and Zhang, L.
\newblock 2017a.
\newblock Cross-domain visual matching via generalized similarity measure and
  feature learning.
\newblock {\em IEEE transactions on pattern analysis and machine intelligence}
  39(6):1089--1102.

\bibitem[\protect\citeauthoryear{Lin \bgroup et al\mbox.\egroup
  }{2017b}]{lin2017active}
Lin, L.; Wang, K.; Meng, D.; Zuo, W.; and Zhang, L.
\newblock 2017b.
\newblock Active self-paced learning for cost-effective and progressive face
  identification.
\newblock {\em IEEE Transactions on Pattern Analysis and Machine Intelligence}.

\bibitem[\protect\citeauthoryear{Lin, Chen, and Yan}{2013}]{lin2013network}
Lin, M.; Chen, Q.; and Yan, S.
\newblock 2013.
\newblock Network in network.
\newblock {\em arXiv preprint arXiv:1312.4400}.

\bibitem[\protect\citeauthoryear{Long, Shelhamer, and
  Darrell}{2015}]{long2015fully}
Long, J.; Shelhamer, E.; and Darrell, T.
\newblock 2015.
\newblock Fully convolutional networks for semantic segmentation.
\newblock In {\em Proceedings of the IEEE Conference on Computer Vision and
  Pattern Recognition},  3431--3440.

\bibitem[\protect\citeauthoryear{Luo, Wu, and Lin}{2017}]{luo2017thinet}
Luo, J.-H.; Wu, J.; and Lin, W.
\newblock 2017.
\newblock Thinet: A filter level pruning method for deep neural network
  compression.
\newblock {\em Proceedings of the IEEE international conference on computer
  vision}.

\bibitem[\protect\citeauthoryear{Molchanov \bgroup et al\mbox.\egroup
  }{2016}]{molchanov2016pruning}
Molchanov, P.; Tyree, S.; Karras, T.; Aila, T.; and Kautz, J.
\newblock 2016.
\newblock Pruning convolutional neural networks for resource efficient transfer
  learning.
\newblock {\em arXiv preprint arXiv:1611.06440}.

\bibitem[\protect\citeauthoryear{Rastegari \bgroup et al\mbox.\egroup
  }{2016}]{rastegari2016xnor}
Rastegari, M.; Ordonez, V.; Redmon, J.; and Farhadi, A.
\newblock 2016.
\newblock Xnor-net: Imagenet classification using binary convolutional neural
  networks.
\newblock In {\em European Conference on Computer Vision},  525--542.
\newblock Springer.

\bibitem[\protect\citeauthoryear{Ren \bgroup et al\mbox.\egroup
  }{2015}]{ren2015faster}
Ren, S.; He, K.; Girshick, R.; and Sun, J.
\newblock 2015.
\newblock Faster r-cnn: Towards real-time object detection with region proposal
  networks.
\newblock In {\em Advances in neural information processing systems},  91--99.

\bibitem[\protect\citeauthoryear{Rigamonti \bgroup et al\mbox.\egroup
  }{2013}]{rigamonti2013learning}
Rigamonti, R.; Sironi, A.; Lepetit, V.; and Fua, P.
\newblock 2013.
\newblock Learning separable filters.
\newblock In {\em Proceedings of the IEEE Conference on Computer Vision and
  Pattern Recognition},  2754--2761.

\bibitem[\protect\citeauthoryear{Russakovsky \bgroup et al\mbox.\egroup
  }{2015}]{russakovsky2015imagenet}
Russakovsky, O.; Deng, J.; Su, H.; Krause, J.; Satheesh, S.; Ma, S.; Huang, Z.;
  Karpathy, A.; Khosla, A.; Bernstein, M.; et~al.
\newblock 2015.
\newblock Imagenet large scale visual recognition challenge.
\newblock {\em International Journal of Computer Vision} 115(3):211--252.

\bibitem[\protect\citeauthoryear{Simonyan and
  Zisserman}{2014}]{simonyan2014very}
Simonyan, K., and Zisserman, A.
\newblock 2014.
\newblock Very deep convolutional networks for large-scale image recognition.
\newblock {\em arXiv preprint arXiv:1409.1556}.

\bibitem[\protect\citeauthoryear{Tai \bgroup et al\mbox.\egroup
  }{2015}]{tai2015convolutional}
Tai, C.; Xiao, T.; Zhang, Y.; Wang, X.; et~al.
\newblock 2015.
\newblock Convolutional neural networks with low-rank regularization.
\newblock {\em arXiv preprint arXiv:1511.06067}.

\bibitem[\protect\citeauthoryear{Wang \bgroup et al\mbox.\egroup
  }{2017}]{wang2017multi}
Wang, Z.; Chen, T.; Li, G.; Xu, R.; and Lin, L.
\newblock 2017.
\newblock Multi-label image recognition by recurrently discovering attentional
  regions.
\newblock In {\em Proceedings of the IEEE Conference on Computer Vision and
  Pattern Recognition},  464--472.

\bibitem[\protect\citeauthoryear{Zhang \bgroup et al\mbox.\egroup
  }{2011}]{zhang2011fsim}
Zhang, L.; Zhang, L.; Mou, X.; and Zhang, D.
\newblock 2011.
\newblock Fsim: A feature similarity index for image quality assessment.
\newblock {\em IEEE transactions on Image Processing} 20(8):2378--2386.

\bibitem[\protect\citeauthoryear{Zhang \bgroup et al\mbox.\egroup
  }{2016}]{zhang2016accelerating}
Zhang, X.; Zou, J.; He, K.; and Sun, J.
\newblock 2016.
\newblock Accelerating very deep convolutional networks for classification and
  detection.
\newblock {\em IEEE transactions on pattern analysis and machine intelligence}
  38(10):1943--1955.

\end{thebibliography}

\end{document}